# A Review of Challenges in Speech-based Conversational AI for Elderly Care

Willemijn KLAASSEN[a], Bram VAN DIJK[b,1], and Marco SPRUIT[b,c]
[a]*Amsterdam Universitair Medische Centra (Amsterdam UMC), The Netherlands*
[b]*Leiden University Medical Center (LUMC), The Netherlands*
[c]*Leiden Institute of Advanced Computer Science (LIACS), The Netherlands*
ORCiD IDs: Willemijn Klaassen https://orcid.org/0009-0001-8324-445,
Bram van Dijk https://orcid.org/0009-0002-9176-1608, Marco Spruit https://orcid.org/0000-0002-9237-221X.

**Abstract.** Artificially intelligent systems optimized for speech conversation are appearing at a fast pace. Such models are interesting from a healthcare perspective, as these voice-controlled assistants may support the elderly and enable remote health monitoring. The bottleneck for efficacy, however, is how well these devices work in practice and how the elderly experience them, but research on this topic is scant. We review elderly use of voice-controlled AI and highlight various user- and technology-centered issues, that need to be considered before effective speech-controlled AI for elderly care can be realized.

**Keywords.** Elderly care, artificial intelligence, conversational AI, geriatrics

## 1. Introduction

The rapid evolution of conversational Artificial Intelligence (AI) has opened new opportunities in healthcare. With an aging population and increasing pressure on the healthcare system, researchers and policymakers are exploring the potential of *speech-based conversational* AI through voice assistants implemented in robots and chatbots [1]. Such technology may enable remote health monitoring, medication reminders and companionship through conversation [2-4], thereby potentially reducing caregiver burden and loneliness for the elderly. Still, conversational AI in healthcare also comes with challenges. Issues include the accessibility, adaptability, and ease of use of these technologies, which may factor negatively in the performance of conversational AI [5]. Worse, AI systems that lack inclusivity in their design may exacerbate existing healthcare disparities, widening the gap in healthcare access and quality between populations. This concise literature review explores current challenges in conversational AI for the elderly. Our results highlight various frequent *user-centered* and *technological* challenges that should inform future work on speech-based AI in elderly care.

## 2. Method

Here we describe the approach for selecting and filtering research, to answer the question what barriers elderly individuals currently face when using voice-assisted AI.

---

[1] Corresponding Author: Bram van Dijk, b.m.a.van_dijk@lumc.nl, Turfmarkt 99, 2511 DP The Hague.

## 2.1. Literature search and study selection

We searched the PubMed database during September-October 2024 for English research articles from 2019 to 2024. We searched using a combination of MeSH and TIAB terms, focusing on keywords relating to the topics *elderly* (e.g. 'geriatrics', 'gerontechnology'), *challenges* (e.g. 'cognition', 'computer literacy'), *artificial intelligence* (e.g. 'speech recognition software', 'natural language processing'), and *evaluation* (e.g. 'biomedical technology assessment', 'usability testing'). This strategy resulted in a total of 571 articles, which were further assessed by the first author.

The free version of the Rayyan software [6] was used to assist with labelling articles for reasons for inclusion and exclusion. First, titles and abstracts were assessed to identify relevant studies, resulting in the exclusion of 512 articles, because of a mismatch in topic, lack of voice-assisted AI, for this review irrelevant outcome measures, or study designs not including an intervention with a voice-controlled device. For the 59 remaining articles, full texts were manually reviewed by the first author and 56 were excluded based on the same criteria, and also articles targeting elderly subpopulations (e.g. with specific diseases). This yielded only 3 articles, and an additional 7 articles were found by forward and backward snowballing, leading to a final set of 10 articles. The full process is depicted in Figure 1.

The final set of articles is small, due to our aim to include research targeting elderly individuals aged over 60 years old, that involved the use of voice-assisted AI as an *intervention*, with relevant outcomes including the performance, and elderly experiences and challenges related to voice-assisted AI usage. Studies that examined voice-assisted AI solely for diagnostic purposes, text-based AI, or caretaker or stakeholder opinions rather than elderly experiences were excluded. Also, we focused on studies from 2019-2024 as new AI tools are appearing at a fast pace, though one study from 2014 was included due to its relevance via backward snowballing.

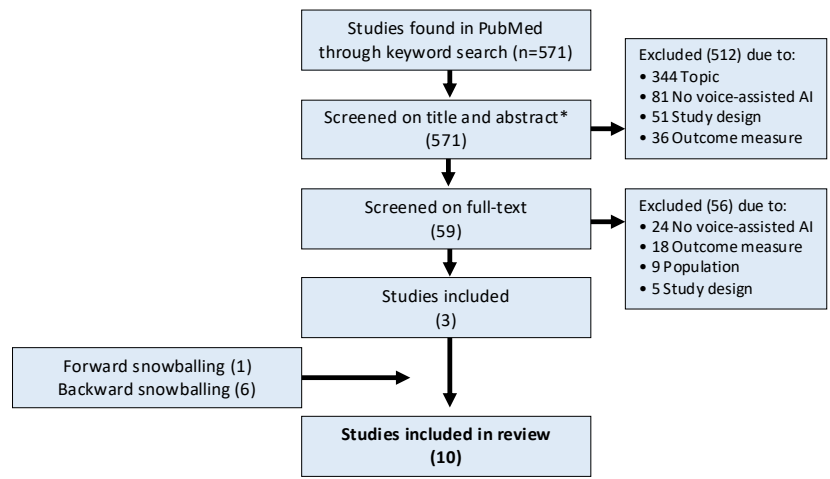

**Figure 1.** Overview of study selection process. Asterisk denotes use of Rayyan software in labelling articles.

*2.2 Study characteristics*

Extracted characteristics of each study included *design, sample characteristics, technological experience*, and *intervention*. Regarding design, all studies included data from qualitative interviews following up on an intervention with a voice-controlled conversational AI device [7-16]. Regarding sample characteristics, in total 241 elderly individuals were included, with most studies having between 10 and 20 participants. Most studies included adults aged 60 years and older [9-16], except for two studies [7,8] which had slightly younger participants (58 and 50 years and older). Most studies included cognitively healthy participants [8-11,14,15]. Concerning technological experience, half of the studies reported no experience [9,12,13,15,16], three studies reported varying levels of experience [7,8,10], and two studies did not directly inquire technological experience [11,14], meaning in most studies the elderly had little technological experience. For intervention, most studies focused on voice-controlled assistants embedded in interfaces such as smart speakers, typically used between one and twelve months [7-11,13,15,16], though one study included short-term use of several hours after which an interview followed [15]. Another study focused on longer term use of a privately owned voice assistant, but specifically on learnability and adoption of the device, thus, we regard it as an intervention and evaluation of the first few months of use.

## 3. Results and discussion

We categorized the challenges we found in elderly use of speech-based conversational AI into *user-centered* and *technological* kinds. This corresponds to the distinction between objective and subjective evaluation of dialogue systems common in the literature [17]. Technological challenges consist of objective problems that directly impact system performance, e.g. software glitches, suboptimal voice recognition, Wi-Fi connectivity problems, and inconsistent responses of the system on the same input. In contrast, user-centered challenges concern the subjective experiences of elderly users, e.g. the usage of wake words, privacy concerns, unfamiliarity with voice assistants, responses to functional errors, and difficulty in question framing. These challenges impact how elderly users interact with speech-based conversational AI. In Table 1 we summarize the challenges and their frequency of occurrence. We elaborate on challenges that occur five times or more in the following two sections.

**Table 1.** Overview of user-centered and technological challenges for the elderly as found in this review.

| User-centered challenges | Freq. | Technological challenges | Freq. |
|---|---|---|---|
| Constructing commands | 10 | Speech recognition | 9 |
| Privacy concerns | 6 | Functional errors | 8 |
| Lack of humanlike interaction | 6 | Complexity of features | 5 |
| Unfamiliarity with technology | 5 | Connectivity issues | 4 |
| Frustration due to error | 4 | Lack of physical controls | 3 |
| Remembering wake word | 4 | Reliability issues | 3 |
| Technological dependency issues | 2 | | |

*3.1 User-centered challenges*

Issues with *constructing commands* were found in all studies [7-16] and encompasses learning how to interact with speech-based conversational AI. Struggles included the elderly using overly complex queries [10,11,13,14,16], pausing too long during speech [15], and remembering key input words [16]. These and similar issues resulted in the device not activating, responding only to the first few words of a query, and ultimately delivering incorrect responses [15]. *Privacy concerns* recurred in six studies [9-10,12,14-16] and emphasized the worry elderly individuals voiced about the confidentiality of voice-assisted devices. Where one study noted participants' unawareness of the device recording capability [14], others found that participants did not believe the recording functionality remained fully private, and displayed worries about devices continuously recording [15,16]. In another study participants limited their use or abandoned the device altogether [14] due to their privacy concerns. A *lack of humanlike interaction* was found in six studies [9,12,13,14-16]; studies noted the issues with a functional and task-oriented dialogue of devices that conflicted with what participants regarded a natural conversation [15,16]. Participants also thought the tone of voice assistants was too rigid and lacked spontaneity [12,14]. Lastly, five studies highlighted *unfamiliarity with the technology* [8,9,11,15,16], which generally involved requiring assistance with setting up the device, misunderstanding how the device operates, or resistance to using technology. The need for assistance [11], printed manuals [16], and sticking to habits [8] revealed that implementing change for the elderly can be particularly difficult.

*3.2 Technological challenges*

*Speech recognition issues* recurred in all but one study [7,9-16]. Problems arose from difficulties in processing speech due to factors as accent, dialect, articulation, stuttering and long pauses between the wake word and further commands [9,10]. Lastly, the elderly often spoke with volume too low for the device to activate [9,11], or voiced lengthy, complex rather than simple commands [14,15], leading to frustration when the voice assistant did not activate. *Functional errors* concerned unexpected system behavior such as inconsistent responses and timeouts, and occurred in eight studies [7-10,13-16]. Systems sometimes returned inconsistent responses to the same questions [7,13] or showed unexplainable behavior such as breakdowns or spontaneous talking [8,14]. Lastly, regarding *complexity of features,* half of the studies highlighted the overwhelming complexity of the conversational AI device [8,11,12,14,16], in terms of complicated responses [14], and too many hard-to-grasp functions [11]. In addition, the absence of conventional buttons sometimes caused confusion in the elderly [8].

*3.3 Discussion*

This review addressed the question what issues the elderly currently face in interacting with state-of-the-art speech-based conversational AI. We focused on experiences of the elderly as opposed to other stakeholders' perspectives, and on interventions meaning that the elderly were exposed to a voice-controlled device for a given period, as opposed to inquiring general opinions about conversational AI. We found many user-centered challenges, indicating a mismatch between elderly and developer/researcher perspectives on private, easy to command, humanlike, and natural conversational AI. Also, the technological challenges we found indicate that state-of-the-art speech-based conversational AI does not handle the nature of elderly speech well and is not sufficiently robust or simple to understand. These findings will benefit developers, researchers, and policymakers in a time where AI is increasingly seen as panacea for many issues in healthcare [18]. Lastly, we note as limitation that in this review, we did not factor the different backgrounds of the elderly in discussion of the results. Future work could further stratify challenges in speech-based conversational AI according to different levels of e.g. technological experience, cognitive functioning, and living situation.